\documentclass[10pt,twocolumn,letterpaper]{article}

\usepackage{cvpr}
\usepackage{times}
\usepackage{epsfig}
\usepackage{graphicx}
\usepackage{amsmath}
\usepackage{amssymb}
\usepackage{lipsum}
\usepackage{multirow}
\usepackage{subfigure}
\usepackage{color}

\usepackage[breaklinks=true,bookmarks=false,colorlinks,linkcolor=red,anchorcolor=blue,citecolor=green]{hyperref}

\cvprfinalcopy 

\def\cvprPaperID{1326} 

\ifcvprfinal\fi
\begin{document}

\title{Foreground-Aware Relation Network for Geospatial Object Segmentation in High Spatial Resolution Remote Sensing Imagery}

\author{Zhuo Zheng\qquad Yanfei Zhong\thanks{Corresponding author. This work was supported by National Key Research and Development
Program of China under Grant No. 2017YFB0504202, National Natural Science Foundation of China under Grant Nos. 41771385, 41801267, and the China Postdoctoral Science Foundation under Grant 2017M622522.}\qquad Junjue Wang\qquad Ailong Ma\\
Wuhan University, Wuhan, China\\
{\tt\small \{zhengzhuo, zhongyanfei, kingdrone, maailong007\}@whu.edu.cn}
}

\maketitle
\thispagestyle{empty}

\begin{abstract}
   Geospatial object segmentation, as a particular semantic segmentation task, always faces with larger-scale variation, larger intra-class variance of background, and foreground-background imbalance in the high spatial resolution (HSR) remote sensing imagery.
   However, general semantic segmentation methods mainly focus on scale variation in the natural scene, with inadequate consideration of the other two problems that usually happen in the large area earth observation scene. 
   In this paper, we argue that the problems lie on the lack of foreground modeling and propose a foreground-aware relation network (FarSeg) from the perspectives of relation-based and optimization-based foreground modeling, to alleviate the above two problems.
   From perspective of relation, FarSeg enhances the discrimination of foreground features via foreground-correlated contexts associated by learning foreground-scene relation.
   Meanwhile, from perspective of optimization, a foreground-aware optimization is proposed to focus on foreground examples and hard examples of background during training for a balanced optimization.
   The experimental results obtained using a large scale dataset suggest that the proposed method is superior to the state-of-the-art general semantic segmentation methods and achieves a better trade-off between speed and accuracy.
   Code has been made available at: \url{https://github.com/Z-Zheng/FarSeg}.
\end{abstract}
\begin{figure}[hbt]
   \begin{center}
      \includegraphics[width=\linewidth]{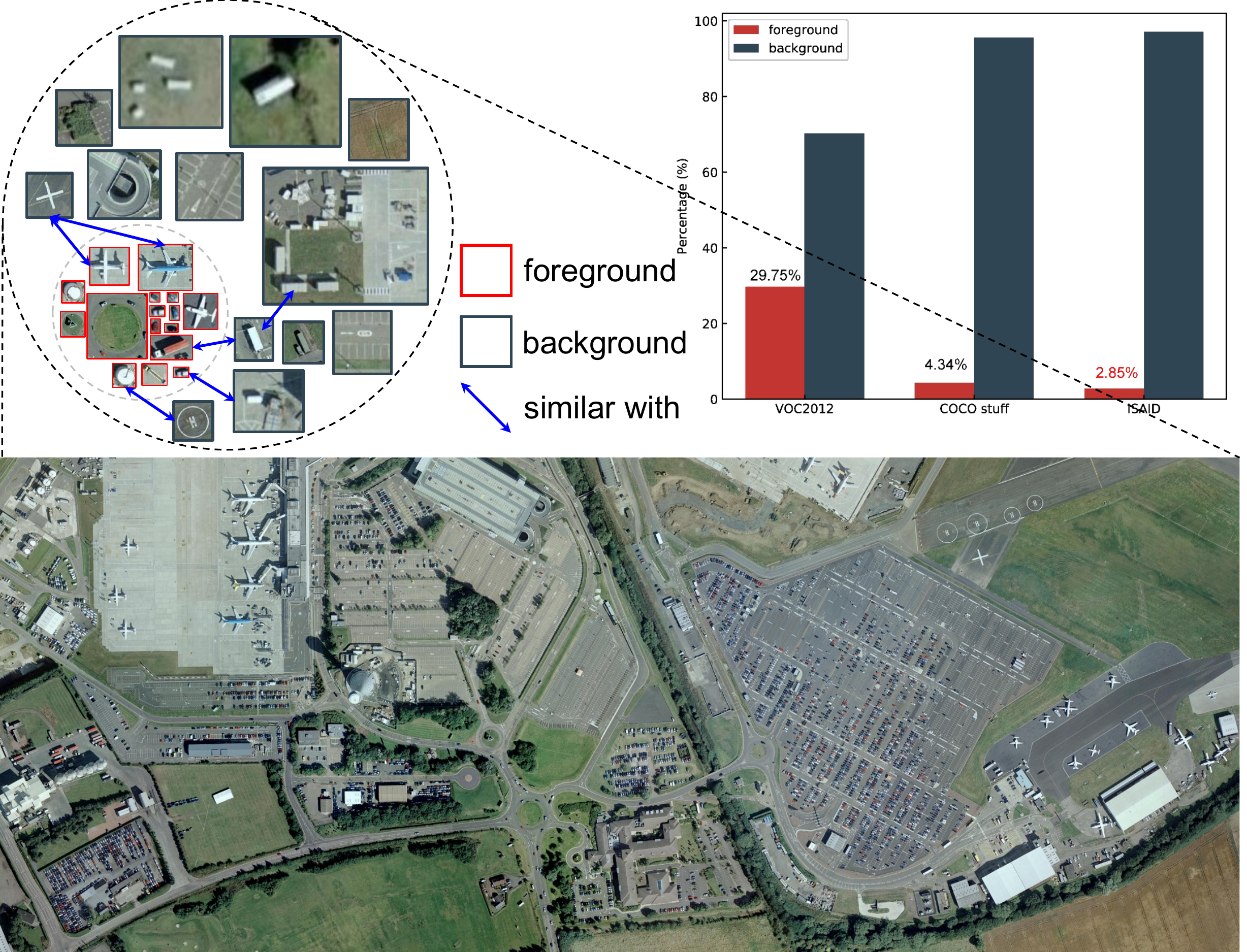}
   \end{center}
   \caption{The main challenges of object segmentation in the HSR remote sensing imagery. (1) larger-scale variation. (2) foreground-background imbalance. (3) intra-class variance of background.}
   \label{fig:coco_voc_isaid}
\end{figure}
\section{Introduction}
High spatial resolution earth observation technique has provided a large number of high spatial resolution (HSR) remote sensing images that can finely describe various geospatial objects, such as ship, vehicle and airplane, etc.
Automatically extracting objects of interest from HSR remote sensing imagery is very helpful for urban management, planing and monitoring \cite{rottensteiner2012isprs,volpi2015semantic,Kampffmeyer_2016_CVPR_Workshops,kemker2018algorithms}.
Geospatial object segmentation, as a significant role in object extraction, can provide semantic and location information for the objects of interest, which belongs to a particular semantic segmentation task with the goal to divide image pixels into two subsets of the foreground objects and the background area.
And meanwhile, it needs to further assign a unique semantic label to each pixel in the foreground object area.

Compared with natural scene, geospatial object segmentation is more challenging in the HSR remote sensing images.
There are three reasons at least:

\textit{1) The object always has larger-scale variation in the HSR remote sensing images \cite{deng2018multi, xia2018dota}.}
This causes the multi-scale problem, which makes it difficult to locate and recognize the object.

\textit{2) The background is much more complex in the HSR remote sensing images \cite{8672899, deng2019learning}}, which causes serious false alarms due to larger intra-class variance.

\textit{3) The foreground ratio is much less than it in the natural images,} as Fig.~\ref{fig:coco_voc_isaid} shows,
which causes foreground-background imbalance problem.

For natural images, the object segmentation task is directly seen as a semantic segmentation task in the computer vision field, the performance of which is mainly limited by the multi-scale problem.
Therefore, current state-of-the-art general semantic segmentation methods focus on scale-aware \cite{chen2016attention} and multi-scale \cite{chen2017deeplab,chen2017rethinking, chen2018encoder,yang2018denseaspp} modeling.
However, for the HSR remote sensing images, false alarms problem and foreground-background imbalance problem are ignored in these general semantic segmentation methods.
We argue that this is because these methods are lack of explicit modeling for the foreground.
This seriously limits the further improvement of object segmentation in the HSR remote sensing images.

In this paper, a foreground-aware relation network (FarSeg) is proposed to tackle aforementioned two problems by exploiting explicitly foreground modeling for more robust object segmentation in the HSR remote sensing imagery.
We explore two perspectives of explicitly foreground modeling: relation-based and optimization-based foreground modeling, and we further propose two modules in the FarSeg: foreground-scene relation module and foreground-aware optimization.
The foreground-scene relation module learns the symbiotic relation between scene and foreground to associate foreground-correlated contexts to enhance the foreground features, thus reducing false alarms.
The foreground-aware optimization focus the model on the foreground by suppressing numerous easy examples in the background to alleviate the foreground-background imbalance problem.

The main contributions of our study are summarized as follows:
\begin{enumerate}
   \item A foreground-aware relation network (FarSeg) is proposed for geospatial object segmentation in HSR remote sensing imagery.
   \item To inherit multi-scale context modeling and learn geospatial scene representation, FarSeg builds a foreground branch based on the feature pyramid network (FPN) and a scene embedding branch upon a shared backbone network, namely multi-branch encoder.
   \item To suppress false alarms, F-S relation module leverages the symbiotic relation between geospatial scene and geospatial objects, to associate foreground-correlated contexts and enhance the discrimination of foreground features. And meanwhile, the background without any contribution is suppressed by this symbiotic relation, thus suppressing false alarms.
   \item To alleviate foreground-background imbalance, F-A optimization is proposed to focus the network on hard examples progressively, thus down-weighting gradient contribution of numerous easy examples in the background, for the foreground-background balanced training.
\end{enumerate}

\section{Related Work}
\label{sec:related_work}

\begin{figure*}
   \begin{center}
      \includegraphics[width=\linewidth]{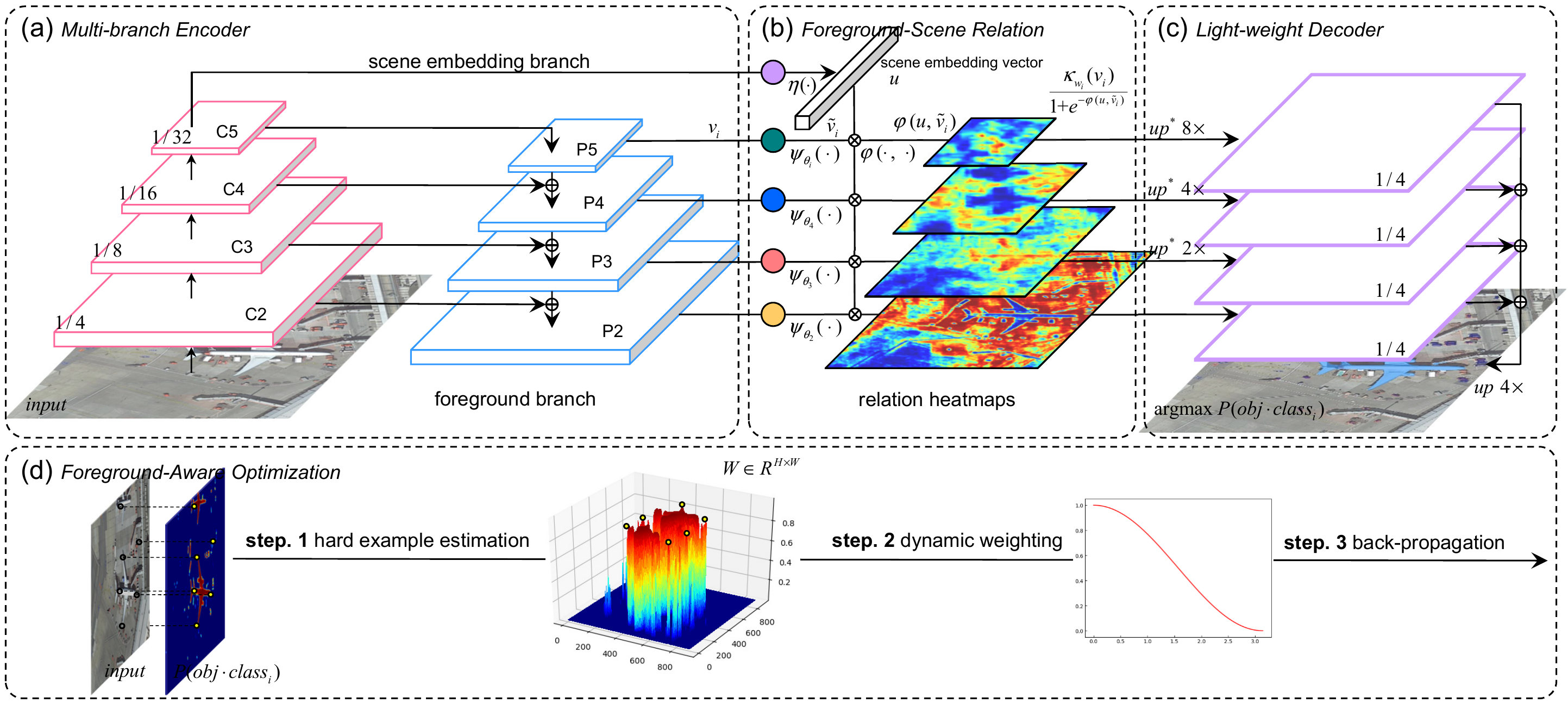}
   \end{center}
   \caption{Overview of FarSeg.
      (a) Multi-branch Encoder for multi-scale object segmentation.
      (b) Foreground-scene relation module.
      (c) Light-weight decoder.
      (d) Foreground-aware optimization. The yellow dots indicate the relative positions of hard example in the raw image, probability map and estimation surface for a simple demonstration.}
   \label{fig:overview}
\end{figure*}

\paragraph{General Semantic Segmentation}

Traditional methods first extract features for each pixel by the handcrafted feature descriptor.
The further promotion of these traditional methods mainly depends on the improvement of handcrafted feature descriptors.
However, designing a feature descriptor is time-consuming and the handcrafted feature is not robust due to limitation of prior knowledge of the expert.

The success of deep learning-based methods lies in solving this problem by learning feature representation from data directly \cite{farabet2012learning}.
Convolutional neural network (CNN), as structured feature representation framework in deep learning, has been explored for semantic segmentation via patch-wise classification \cite{ciresan2012deep,farabet2012learning,hariharan2014simultaneous,gupta2014learning,pinheiro2014recurrent}.
However, patch-wise fashion limits the spatial context modeling and brings redundant computation on overlapped areas between patches.
To solve this problem, fully convolutional network (FCN) \cite{long2015fully} was proposed, which directly outputs the pixel-wise prediction from the input with arbitrary size via the in-network upsampling layer.
FCN was the first pixels-to-pixels semantic segmentation method and was end-to-end trained.

To further exploit spatial context for semantic segmentation, deeplab v1 \cite{chen2014semantic} utilized atrous convolution to enlarge receptive field of the CNN for wider spatial context modeling.
And a dense conditional random field (CRF) was used as a postprocess to smooth the prediction.

To learn multi-scale feature representation, atrous spatial pyramidal pooling (ASPP) \cite{chen2017deeplab} and pyramid pooling module (PPM) \cite{zhao2017pyramid} were proposed.
ASPP utilized multiple atrous convolutions with different atrous rate to extract features with the different receptive field, while PPM generated pyramidal feature maps via pyramid pooling \cite{he2015spatial}.
The image-level features and batch normalization were embedded into ASPP for further improvement of accuracy in deeplab v3 \cite{chen2017rethinking}.
DenseASPP \cite{yang2018denseaspp} further enhanced multi-scale feature representation via densely connected ASPP to make the multi-scale features covering larger and denser scale range.
However, these methods failed to extract fine details of the object, such as the edge.

U-Net \cite{ronneberger2015u} and SegNet \cite{badrinarayanan2017segnet} utilized a new ``encoder-decoder'' network architecture, which reused the shallow features with high spatial resolution to enhance the deep features with strong semantics on spatial detail.
RefineNet \cite{lin2017refinenet} proposed a multi-path refinement network to progressively recover the spatial detail of deep features for better accuracy and visual performance.
Deeplab v3+ also adopted ``encoder-decoder'' framework to further improve performance via a more powerful backbone Xception \cite{chollet2017xception} and a light-weight decoder to recover the spatial resolution of features with a small overhead.

These general semantic segmentation methods mainly focus on multi-scale context modeling, ignoring the special issues in the HSR remote sensing imagery, such as false alarms and foreground-background imbalance.
This causes that these methods are lack of explicit modeling for the foreground.
Therefore, a foreground-aware method is needed for object segmentation in the HSR remote sensing imagery.
\paragraph{Semantic Segmentation in Remote Sensing Community}

There are a lot of applications using semantic segmentation technique in the remote sensing community, such as land use and land cover (LULC) classification \cite{zhang2018object,huang2018urban,zhang2019joint}, building extraction \cite{yuan2017learning, ji2018fully, xu2018building,dickenson2018rotated}, road extraction \cite{liang2019convolutional,cheng2017automatic,bastani2018roadtracer,lu2019multi,batra2019improved}, vehicle detection \cite{mou2018vehicle}, etc.
The main methodologies follow general semantic segmentation, but for special application scenario (e.g. road or building), there were many improved techniques \cite{bastani2018roadtracer, dickenson2018rotated, batra2019improved} for its application scenario.

However, these methods mainly focus on the improvement under the special application scenario, ignoring the consideration of common issues for object segmentation in the HSR remote sensing imagery, such as false alarms problem and foreground-background imbalance problem, especially for large scale HSR remote sensing imagery.
Hence, we propose a foreground-aware relation network (FarSeg) to tackle these problems.

\section{Foreground-Aware Relation Network}
\label{sec:method}
To explicit model the foreground for object segmentation in the HSR remote sensing imagery, we propose a foreground-aware relation network (FarSeg), as shown in Fig.~\ref{fig:overview}.
The proposed FarSeg consists of a variant of feature pyramid network (FPN), foreground-scene (F-S) relation module, light-weight decoder and foreground-aware (F-A) optimization.
FPN is responsible for multi-scale object segmentation.
In the F-S relation module, we first formulate false alarms problem as a problem of lacking discriminative information in the foreground, and then introduce the latent scene semantics and F-S relation to improve the discrimination of foreground features.
The light-weight decoder is simply designed to recover the spatial resolution of semantic features.
To make the network focus on foreground during training, the F-A optimization is proposed to alleviate foreground-background imbalance problem.

\subsection{Multi-Branch Encoder}
Multi-branch encoder is made up of a foreground branch and a scene embedding branch.
As shown in Fig.~\ref{fig:overview} (a), these branches are built upon a backbone network.
In the proposed method, ResNets \cite{he2016deep} are chosen as the backbone network for basic feature extraction.
\{$C_i | i=2,3,4,5$\} denotes the set of feature maps extracted from ResNets, where the feature map $C_i$ has a output stride of $2^{i}$ pixels with respect to the input image.
Similar to the original FPN, the top-down pathway and lateral connections are used to generated pyramidal feature maps \{$P_i | i=2,3,4,5$\} with a same number of channels $d$.
We formulate this procedure as follows:
\begin{equation}
   P_i = \zeta(C_i) + \Gamma(P_{i+1}), i=2,3,4,5
\end{equation}
where $\zeta$ denotes the lateral connection implemented by a learnable 1$\times$1 convolutional layer and $\Gamma$ denotes a nearest neighbor upsampling with a scale factor of 2.
By this top-down pathway and lateral connections, the feature maps can be enhanced with high spatial detail from shallow layers and strong semantics from deep layers, which is helpful to recover the detail of objects and multi-scale context modeling.
Apart from the pyramidal feature maps $v_i$, a extra branch is attached on $C_5$ to generate a geospatial scene feature $C_6$ via global context aggregation.
For simplicity, we use global average pooling as the aggregation function.
$C_6$ is used to model the relation between geospatial scene and foreground, which is illustrated in the Section \ref{sec:relation}.

\subsection{Foreground-Scene Relation Module}
\label{sec:relation}
\begin{figure}[hbt]
   \begin{center}
      \includegraphics[width=0.9\linewidth]{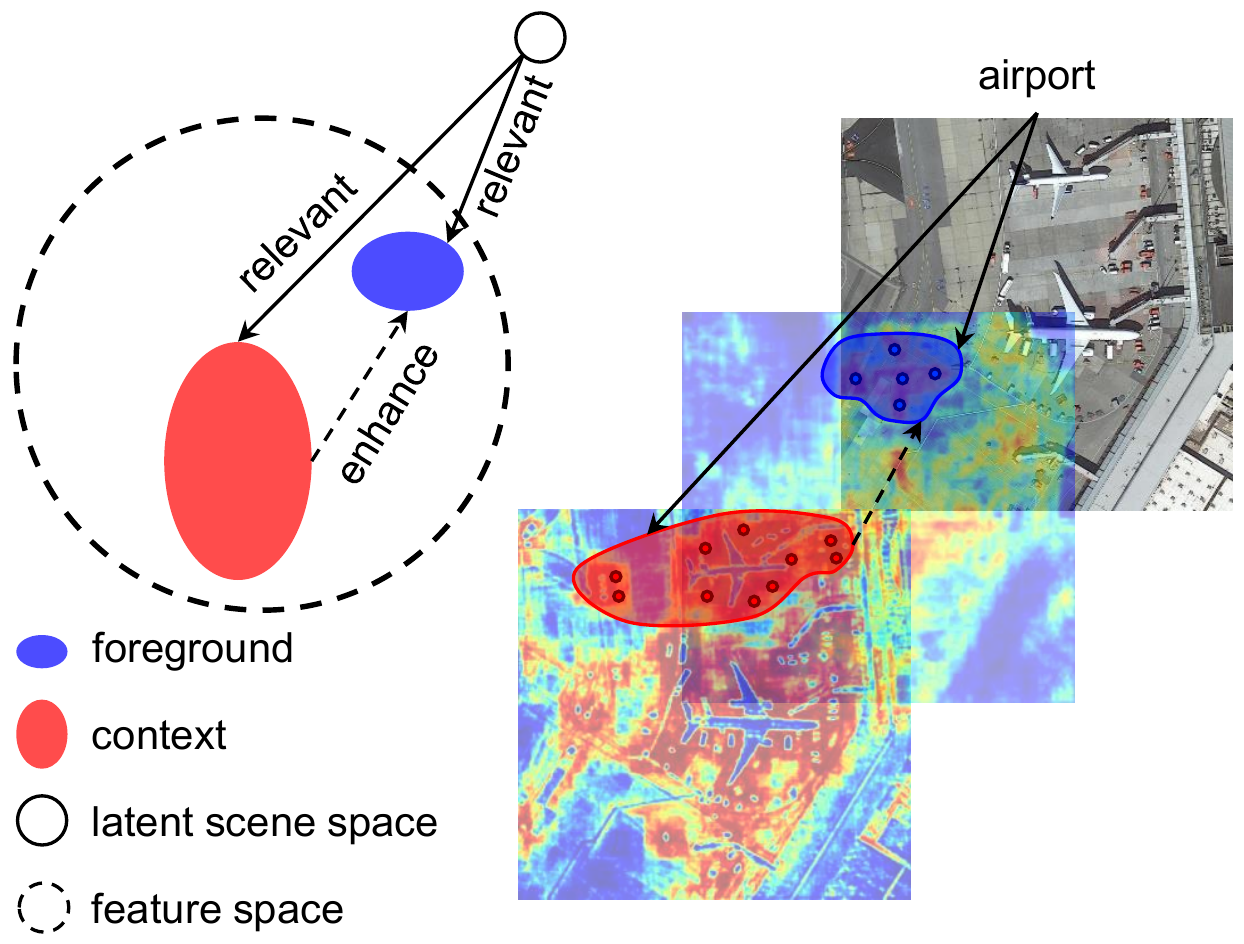}
   \end{center}
   \caption{Concept of F-S relation.
      The foreground features are associated with relevant context features by their collaborative latent geospatial scene space. Meanwhile, the relevant context features are utilized to enhance the discrimination of the foreground features.}
   \label{fig:relation_concept}
\end{figure}

The background is much more complex in the HSR remote sensing imagery.
It means that there is larger intra-class variance in the background, which causes the false alarms problem.
To alleviate this problem, foreground-scene (F-S) relation module is proposed to improve the discrimination of foreground features by associating geospatial scene-relevant context.
The main idea is shown in Fig.~\ref{fig:relation_concept}.
F-S relation module first explicitly models the relation between foreground and geospatial scene and use latent geospatial scene to associate the foreground and relevant context.
And then the relation is used to enhance the input feature maps to increase the disparity  between foreground features and background features, thereby improving the discrimination of foreground features.

As shown in Fig.~\ref{fig:overview} (b), for the pyramidal feature map $v_i$, F-S relation module will produce a new feature map $z_i$.
The feature map $z_i$ is obtained by re-encoding $v_i$ and then re-weighting it using the relation map $r_i$.
The relation map $r_i$ is the similarity matrix between geospatial scene representation and foreground representation.
To align these two feature representations into a shared manifold $R^{d_u}$, there are two projection functions needed to learn for geospatial scene and foreground, respectively.
$\tilde{v}_i$ is the feature map $v_i$ transformed by the scale-aware projection function $\psi_{\theta_i}(\cdot): R^{d\times H\times W} \mapsto R^{d_u\times H\times W}$
, as shown in Eqn.~\ref{eqn:scale_aware_encoder}.
\begin{equation}
   \label{eqn:scale_aware_encoder}
   \tilde{v}_i = \psi_{\theta_i}(v_i)
\end{equation}
where $\theta_i$ denotes the learnable parameters of $\psi_{\theta_i}(\cdot)$.
We adopt a simple form of $\psi_{\theta_i}(\cdot)$ which is just implemented by 1$\times$1 convolutional layer followed by batch normalization and ReLU in order.

\begin{figure}[hbt]
   \begin{center}
      \includegraphics[width=\linewidth]{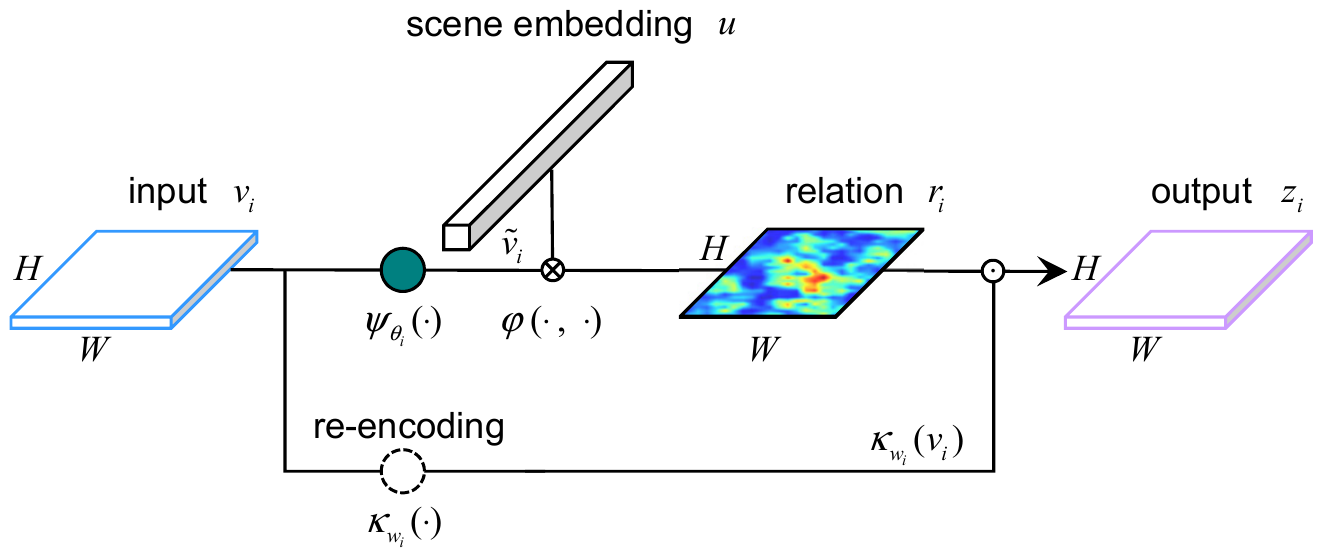}
   \end{center}
   \caption{The computation detail of relation modeling for the pyramid level $i$ in the F-S relation module. The input and output have the same spatial size.}
   \label{fig:relation_module}
\end{figure}

To compute the relation map $r_i$, a 1-D scene embedding vector $u \in R^{d_u}$ is needed to interact with the foreground feature maps $\tilde{v}_i$ in the shared manifold.
The scene embedding vector $u$ is computed by applying $\eta(\cdot)$ on $C_6$, as shown Eqn.~\ref{eqn:u}.
\begin{equation}
   \label{eqn:u}
   u = \eta(C_6)
\end{equation}
where $\eta$ denotes a projection function for geospatial scene representation and it is implemented by a learnable 1$\times$1 convolutional layer with output channels of $d_u$.
The scene embedding vector $u$ is shared for each pyramid because the latent geospatial scene semantics is scale-invariant cross all pyramids.
Hence, the relation map $r_i$ can be naturally obtained by Eqn.~\ref{eqn:r}.
\begin{equation}
   \label{eqn:r}
   r_i = \varphi(u, \tilde{v}_i) = u\odot \tilde{v}_i
\end{equation}
where $\varphi$ denotes the similar estimation function and it is implemented by pointwise inner product for simplicity and efficient computational complexity.

For each pyramid level, the process detail of the relation modeling is illustrated in Fig.~\ref{fig:relation_module} and relation enhanced foreground feature maps $z_i$ is computed as follows:
\begin{equation}
   \label{eqn:z}
   z_i = \frac{1}{1+\exp(-r_i)} \cdot \kappa_{w_i}(v_i)
\end{equation}
where $\kappa_{w_i}(\cdot)$ is the encoder with learnable parameters $w_i$ for input feature maps $v_i$.
The encoder is designed to introduce a extra non-linear unit to avoid feature degradation since the weighting operation is a linear function.
Therefore, we adopt a simple form of this encoder, which implemented by a 1$\times$1 convolutional layer followed by batch normalization and ReLU for high efficiency of parameters and computation.
The item including $r_i$ of Eqn.~\ref{eqn:z} is used to weight the re-encoded feature maps,
which is the normalized relation map using the sigmoid gate function based on a simple self-gating mechanism \cite{hu2018squeeze}.

\subsection{Light-weight decoder}
\begin{figure}[hbt]
   \begin{center}
      \includegraphics[width=\linewidth]{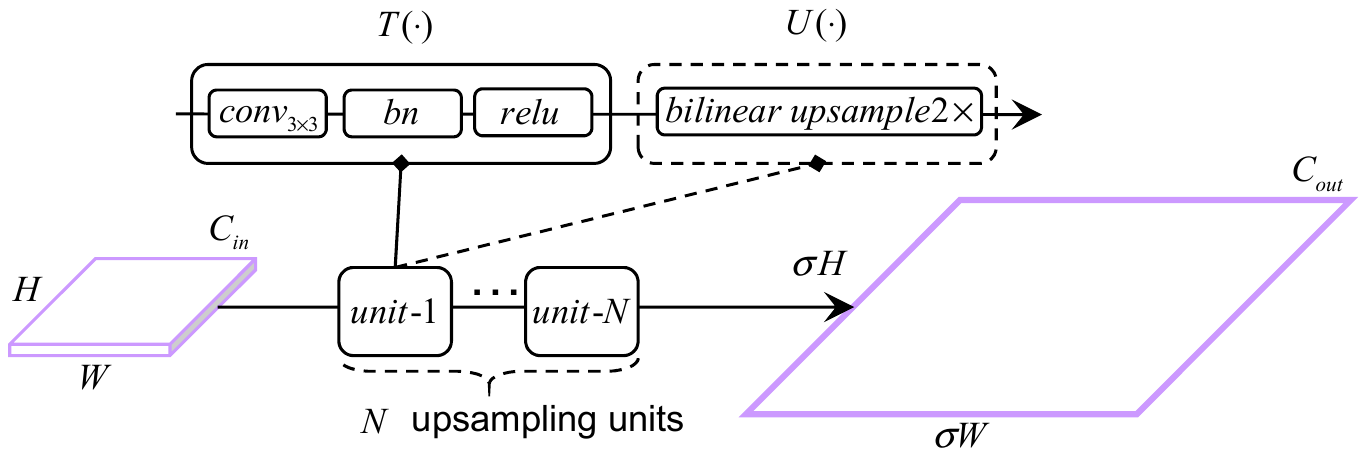}
   \end{center}
   \caption{Abstract architecture of the light-weight decoder for each pyramid level.}
   \label{fig:decoder}
\end{figure}

The light-weight decoder is designed to recover the spatial resolution of relation enhanced semantic feature maps from F-S relation module in a light-weight fashion.
The detailed architecture of the light-weight decoder is illustrated in Fig.~\ref{fig:decoder}.

Given the pyramid feature maps $z_i \in R^{C_{in}\times H\times W}$  from F-S relation module, the upsampled feature maps $z_i' \in R^{C_{out}\times \sigma H\times \sigma W}$ is computed via the light-weight decoder.
The light-weight decoder is stacked by many upsampling units.
The upsampling unit is made up of a channel transformation $T(\cdot)$ and an optional 2$\times$ upsampling operation $U(\cdot)$, which only includes $T(\cdot)$ if the scale factor $\sigma = 1$.
Hence, the light-weight decoder for pyramid level $i$ can be simply formulated as:
\begin{equation}
   \label{eqn:zi}
   z_i'=\left\{
   \begin{aligned}
      \underbrace{U\circ T}_{N}(z_i) & , & N > 0, \\
      T(z_i)                         & , & N = 0.
   \end{aligned}
   \right.
\end{equation}
where $N$ denotes the number of upsampling units and $N = i - 2$.

$T(\cdot)$ is implemented by a 3$\times$3 convolutional layer followed by batch normalization and ReLU.
$U(\cdot)$ is the bilinear upsampling with a scale factor of 2.
The total upsampling scale $\sigma$ is equal to $2^N$ because the output stride is 4.

To aggregate upsampled feature maps from each pyramid, the point-wise mean operation followed by 1$\times$1 convolutional layer is adopted for computation and parameter efficiency.
And a 4$\times$ bilinear upsampling is used to produce the final class probability map of the same size as the input image.

\subsection{Foreground-Aware Optimization}
The foreground-background imbalance problem usually causes the fact that background examples dominate the gradients during training.
However, only the hard part of the background examples is valuable for optimization in the late period of training, where the hard examples are much less than easy examples in the background.
Motivated by this, the foreground-aware optimization is proposed to make the network focus on foreground and hard examples in the background for a balanced optimization.
The foreground-aware optimization includes three steps: hard example estimation, dynamic weighting and back-propagation, as shown in Fig.~\ref{fig:overview} (d).

\textbf{hard example estimation.}
This step is used to obtain the weights reflecting the hard degree of examples to adjust the distribution of pixel-wise loss.
That the example is harder means that its weight is larger.
Motivated by focal loss \cite{lin2017focal}, we adopt $(1 - p)^{\gamma}$ as weight to estimate hard examples, where $p \in [0, 1]$ is the predicted probability by the network and $\gamma$ is the focusing factor.
This formulation was used in object detection, but for the pixel-level task with foreground-background imbalance, we only expect to adjust the loss distribution without change of sum for avoiding gradient vanishing.
Therefore, we generalize it for object segmentation in the HSR remote sensing imagery by introducing a normalization constant $Z$ that guarantees $\sum{l(p_i, y_i)} = \frac{1}{Z}\sum{(1-p_i)^\gamma l(p_i, y_i)}$, where $l(p_i, y_i)$ denotes the cross entropy loss of $i$-th pixel computed by predicted probability $p_i$ and its ground truth $y_i$.
Hence, for the loss of each pixel, it has a weight $\frac{1}{Z}(1-p_i)^\gamma$.

\textbf{dynamic weighting.}
The hard example estimation relies on the discrimination of the model.
However, the discrimination is unconfident in the initial period of training, which makes the hard example estimation unconfident.
If this unconfident hard example weights are used, the model training will be unstable, influencing the converged performance.
To solve this problem, we propose a dynamic weighting strategy based on an annealing function.
We design three annealing functions as the candidates, as Table ~\ref{tab:annealing_func} lists.
Given the cross entropy loss $l(p_i, y_i)$, the dynamic weighted loss is formulate as:
\begin{equation}
   l'(p_i, y_i) = [\frac{1}{Z}(1 - p_i)^{\gamma} + \zeta(t)(1 - \frac{1}{Z}(1 - p_i)^{\gamma})]\cdot l(p_i, y_i)
\end{equation}
where $\zeta(\cdot)$ denotes an annealing function with respect to current training step $t$ and $\zeta(t) \in [0, 1]$ is a monotonically decreasing function.
By this way, the focus of loss distribution can progressively move on hard examples with the increase of the confidence of hard example estimation.

\begin{table}[]
   \caption{Candidates of annealing functions.
      \label{tab:annealing_func}}
   \centering
   \renewcommand{\arraystretch}{1.5}
   \resizebox{\linewidth}{!}{
      \begin{tabular}{lll}
         \hline
         Annealing function & \multicolumn{1}{c}{Formula}                                    & Hyperparameter                     \\ \hline
         Linear             & $\zeta(t) = 1 - \frac{t}{annealing\_step} $                    & $annealing\_step$                  \\
         Poly               & $\zeta(t) = (1 - \frac{t}{annealing\_step})^{decay\_factor} $  & $annealing\_step$, $decay\_factor$ \\
         Cosine             & $\zeta(t) = 0.5 * (1 + \cos{(\frac{t}{annealing\_step}\pi)}) $ & $annealing\_step$                  \\ \hline
      \end{tabular}}
\end{table}

\section{Experiments}
\label{sec:exp}

\subsection{Experimental setting}

\paragraph{Dataset.}

iSAID \cite{waqas2019isaid} dataset consists of 2,806 HSR remote sensing images.
These images were collected from multiple sensors and platforms with multiple resolutions.
The original image sizes range from $\sim$ 800 $\times$ 800 pixels to $\sim$ 4000 $\times$ 13000 pixels.
The iSAID dataset provides 655,451 instances annotations over 15 categories\footnote{The categories are defined as: ship (Ship), storage tank (ST), baseball diamond (BD), tennis court (TC), basketball court (BC), ground field track (GTF), bridge (Bridge), large vehicle (LV), small vehicle (SV), helicopter (HC), swimming pool (SP), roundabout (RA), soccerball field (SBF), plane (Plane), harbor (Harbor).}
of the object, which is the largest dataset for instance segmentation in the HSR remote sensing imagery.
The predefined training set contains 1,411 images, while validation (\textit{val}) set contains 458 images and test set has 937 images.
In this work, we only use semantic mask annotations for object segmentation.
And we use the predefined training set to train models and evaluate on the validation set.
Because the test set is unavailable.

\paragraph{Implementation detail.}
The backbone used in FarSeg was ResNet-50 for all the experiments, which was pretrained on ImageNet \cite{deng2009imagenet}.
The channels $d$ in FPN was set to 256 and the dimension of shared manifold $d_u$ in F-S relation module was set to 256 if not specified.
The default focusing factor $\gamma$ in F-A optimization was 2.
For hyperparameters introduced by F-A optimization, $annealing\_step$ was set to 10k and $decay\_factor$ was set to 0.9 for the poly annealing function.
For all the experiments, these models were trained for 60k iterations with a ``poly'' learning rate policy, where the initial learning rate was set to 0.007 and multiplied by $(1 - \frac{step}{max\_step})^{power}$ with $power = 0.9$.
We used synchronized SGD over 2 GPUs with a total of 8 images per mini-batch (4 images per GPU), weight decay of 0.0001 and momentum of 0.9.
The synchronized batch normalization was used for cross-gpu communication of statistic in the batch normalization layer.
For data augmentation, horizontal and vertical flip, rotation of $90\cdot k~(k=1,2,3)$ degree were adopted during training.
For extra data preprocessing, we crop the image into a fixed size of (896, 896) using a sliding window striding 512 pixels.

\paragraph{Evaluation metric.}
Following the common practice \cite{everingham2015pascal, lin2014microsoft}, we used the mean intersection over union (mIoU) as the main metric for object segmentation to evaluate the proposed method.

\subsection{Comparison to General methods}
To evaluate the FarSeg, we conduct comprehensive experiments on a larger scale HSR remote sensing images dataset.
We compared FarSeg with several CNN-based methods from classical to state-of-the-art, including U-Net \cite{ronneberger2015u}, FCN-8s \cite{long2015fully}, DenseASPP \cite{yang2018denseaspp}, Deeplab v3 \cite{chen2017rethinking}, Semantic FPN \cite{kirillov2019panoptic}, Deeplab v3+ \cite{chen2018encoder}, RefineNet \cite{lin2017refinenet}, PSPNet \cite{zhao2017pyramid}.
The quantitative results listed in Table~\ref{tab:isaid_sota} suggest that FarSeg outperforms other methods in HSR scenario.

\begin{figure}[hbt]
   \begin{center}
      \includegraphics[width=0.8\linewidth]{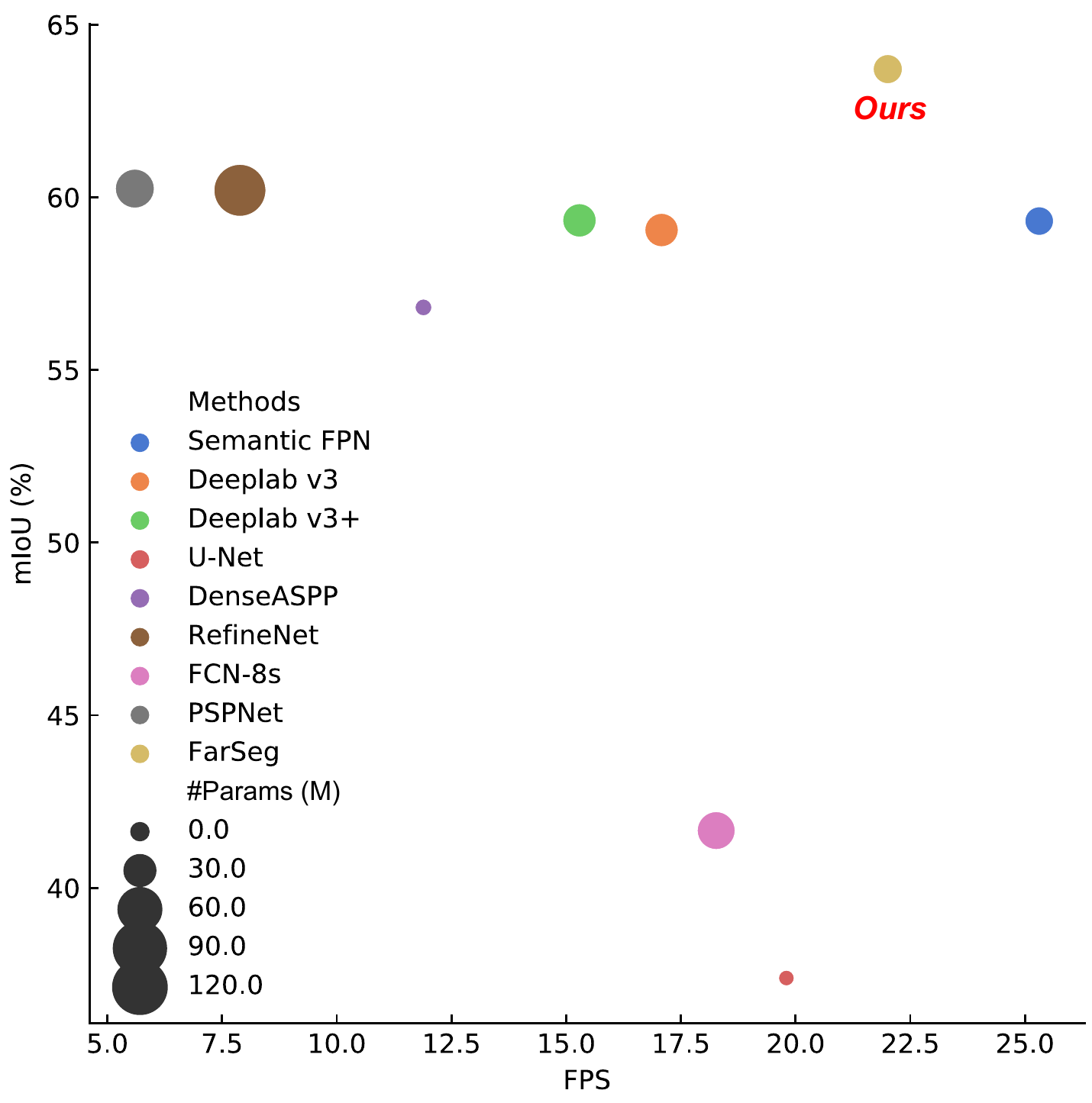}
   \end{center}
   \caption{Speed (FPS) versus accuracy (mIoU) on iSAID $val$ set. The radius of circles represents the number of parameters.}
   \label{fig:speed_accuracy}
\end{figure}

Fig.~\ref{fig:speed_accuracy} shows the trade-off between speed and accuracy.
It indicates that FarSeg achieves a better trade-off between speed and accuracy, which benefits from the light-weight and effective module design.

\begin{table*}[hbt]
   \caption{Object segmentation mIoU (\%) on iSAID \textit{val} set.
      The bold values in each column means the best entries.
      \label{tab:isaid_sota}}
   \centering
   \renewcommand{\arraystretch}{1.5}
   \resizebox{\linewidth}{!}{
      \begin{tabular}{l|l|c|ccccccccccccccc}
         \hline
         \multirow{2}{*}{Method}              & \multirow{2}{*}{backbone} & \multirow{2}{*}{mIoU (\%)} & \multicolumn{15}{c}{IoU per category (\%)}                                                                                                                                                                                                                                               \\ \cline{4-18}
                                              &                           &                            & Ship                                       & ST             & BD             & TC             & BC             & GTF            & Bridge         & LV             & SV             & HC             & SP             & RA             & SBF            & Plane          & Harbor         \\ \hline
         U-Net \cite{ronneberger2015u}        & -                         & 37.39                      & 49.0                                       & 0              & 6.51           & 78.60          & 22.89          & 5.52           & 7.48           & 49.89          & 35.62          & 0              & 38.03          & 46.49          & 9.67           & 74.74          & 45.64          \\ \hline
         FCN-8s \cite{long2015fully}          & VGG-16                    & 41.66                      & 51.74                                      & 22.91          & 26.44          & 74.81          & 30.24          & 27.85          & 8.17           & 49.35          & 37.05          & 0              & 30.74          & 51.91          & 52.07          & 62.90          & 42.02          \\ \hline
         DenseASPP \cite{yang2018denseaspp}   & DenseNet-121              & 56.81                      & 61.15                                      & 50.05          & 67.54          & 86.09          & 56.56          & 52.28          & 29.61          & 57.10          & 38.44          & 0              & 43.26          & 64.80          & 74.10          & 78.12          & 51.09          \\ \hline
         Deeplab v3 \cite{chen2017rethinking} & ResNet-50                 & 59.05                      & 59.74                                      & 50.49          & 76.98          & 84.21          & 57.92          & 59.57          & 32.88          & 54.80          & 33.75          & 31.29          & 44.74          & 66.03          & 72.13          & 75.84          & 45.68          \\ \hline
         Semantic FPN \cite{kirillov2019panoptic}    & ResNet-50                 & 59.31                      & 63.68                                      & 59.49          & 71.75          & \textbf{86.61} & 57.78          & 51.64          & 33.99          & 59.15          & 45.14          & 0              & 46.42          & 68.71          & 73.58          & 80.83          & 51.27          \\ \hline
         Deeplab v3+ \cite{chen2018encoder}   & ResNet-50                 & 59.33                      & 59.02                                      & 55.15          & 75.94          & 84.18          & 58.52          & 59.24          & 32.11          & 54.54          & 33.79          & 31.14          & 44.24          & 67.51          & 73.78          & 75.70          & 45.76          \\ \hline
         RefineNet \cite{lin2017refinenet}    & ResNet-50                 & 60.20                      & 63.80                                      & 58.56          & 72.31          & 85.28          & 61.09          & 52.78          & 32.63          & 58.23          & 42.36          & 22.98          & 43.40          & 65.63          & \textbf{74.42} & 79.89          & 51.10          \\ \hline
         PSPNet \cite{zhao2017pyramid}        & ResNet-50                 & 60.25                      & 65.2                                       & 52.1           & 75.7           & 85.57          & 61.12          & \textbf{60.15} & 32.46          & 58.03          & 42.96          & 10.89          & 46.78          & 68.6           & 71.9           & 79.5           & \textbf{54.26} \\ \hline\hline
         FarSeg                               & ResNet-50                 & \textbf{63.71}             & \textbf{65.38}                             & \textbf{61.80} & \textbf{77.73} & 86.35          & \textbf{62.08} & 56.70          & \textbf{36.70} & \textbf{60.59} & \textbf{46.34} & \textbf{35.82} & \textbf{51.21} & \textbf{71.35} & 72.53          & \textbf{82.03} & 53.91          \\ \hline
      \end{tabular}
   }
\end{table*}

\begin{table}[]
   \caption{Object segmentation mIoU (\%) on iSAID \textit{val} set. Starting from Baseline, the proposed modules are gradually added in the proposed FarSeg for the module analysis.
      \label{tab:module_ablation}}
   \centering
   \renewcommand{\arraystretch}{1.5}
   \resizebox{\linewidth}{!}{
      \begin{tabular}{l|ccc|cc}
         \hline
         Method                                             & \multicolumn{1}{l}{F-S Relation} & \multicolumn{1}{l}{Scale-aware Proj.} & \multicolumn{1}{l|}{F-A \textbf{Opt.}} & \multicolumn{1}{l}{mIoU(\%)} & $\Delta$\#params(M) \\ \hline
         (a) Baseline                                       & -                                & -                                     & -                                      & 59.31                        & 0                   \\ \hline

         (b) Baseline w/ F-S Relation                       & $\checkmark$                     &                                       &                                        & 60.42                        & 1.12                \\
         (c) Baseline w/ F-S Relation and Scale-aware Proj. & $\checkmark$                     & $\checkmark$                          &                                        & 60.49                        & 2.89                \\
         (d) Baseline w/ F-A Opt.                           &                                  &                                       & $\checkmark$                           & 61.51                        & 0                   \\
         (e) Baseline w/ F-S Relation and F-A Opt.          & $\checkmark$                     &                                       & $\checkmark$                           & 63.21                        & 1.12                \\
         (f) FarSeg                                         & $\checkmark$                     & $\checkmark$                          & $\checkmark$                           & 63.71                        & 2.89                \\ \hline
      \end{tabular}}
\end{table}

\begin{table}[]
   \caption{Foreground-aware optimization module analysis.
      \label{tab:opt_ablation}}
   \centering
   \renewcommand{\arraystretch}{1.5}
   \resizebox{\linewidth}{!}{
      \begin{tabular}{ll|cc|c}
         \hline
         \multicolumn{2}{l|}{Method}                                & \multicolumn{1}{l}{Normalization} & \multicolumn{1}{l}{Annealing function} & \multicolumn{1}{|l}{mIoU(\%)}         \\ \hline
         \multicolumn{2}{l|}{(a) FarSeg w/o F-A \textbf{Opt.}}      & -                                 & -                                      & 60.49                                 \\ \hline
         \multicolumn{2}{l|}{(b) Loss weighted with $(1-p)^\gamma$} &                                   &                                        & 56.44                                 \\
         \multicolumn{2}{l|}{(c) + \textbf{Norm.}}                  & $\checkmark$                      &                                        & 62.98                                 \\
         (d) + \textbf{Norm.}                                       & + Linear Annealing                & $\checkmark$                           & Linear                        & 63.18 \\

         (e) + \textbf{Norm.}                                       & + Poly Annealing                  & $\checkmark$                           & Poly                          & 63.52 \\
         (f) + \textbf{Norm.}                                       & + Cosine Annealing                & $\checkmark$                           & Cosine                        & 63.71 \\
         \hline
      \end{tabular}}
\end{table}

\subsection{Ablation Study}
In this section, we conduct comprehensive experiments to analyze the proposed modules and many important hyper-parameters in FarSeg.
The baseline is composed of a FPN and a light-decoder, optimizing cross entropy loss.
The mIoU is evaluated on iSAID \textit{val} set with the same experimental settings if not specified.

\subsubsection{Foreground-Scene Relation Module}

\paragraph{The effect of F-S relation module.}
Table~\ref{tab:module_ablation} (b) and (c) show the ablation results of adding F-S relation based on baseline method (Table~\ref{tab:module_ablation} (a)).
F-S relation modules (w/o and w/ scale-aware projection) brings 1.11\% and 1.18\% performance gains in mIoU, respectively.
$\Delta$\#params denotes the extra parameters introduced by the corresponding module.
It indicates that F-S relation modules are parameter efficient with only 2.89 M and 1.12 M, where relative increments of parameters are $\sim$ 10\% and $\sim$ 4\%, respectively.
This suggests that the performance gain not only comes from the gain of parameters, but also results from F-S relation design of using geospatial scene feature associates the relevant context features to enhance the foreground features.

\paragraph{Scale-aware projection for scene embedding.}
The projection function $\eta$ is used for geospatial scene representation in F-S relation module.
We explore whether the scale-aware projection function $\eta$ is needed for each pyramid level.
The results of Table~\ref{tab:module_ablation} (b)/(c) and (e)/(f) suggest that scale-aware projection function performs better.
With F-A optimization, the gain in mIoU from scale-aware projection is larger.
It indicates that geospatial scene representation is related to scale and foregrounds.

\begin{figure*}
   \centering
   \subfigure[Image]{
      \begin{minipage}[b]{0.15\linewidth}
         \includegraphics[width=\linewidth]{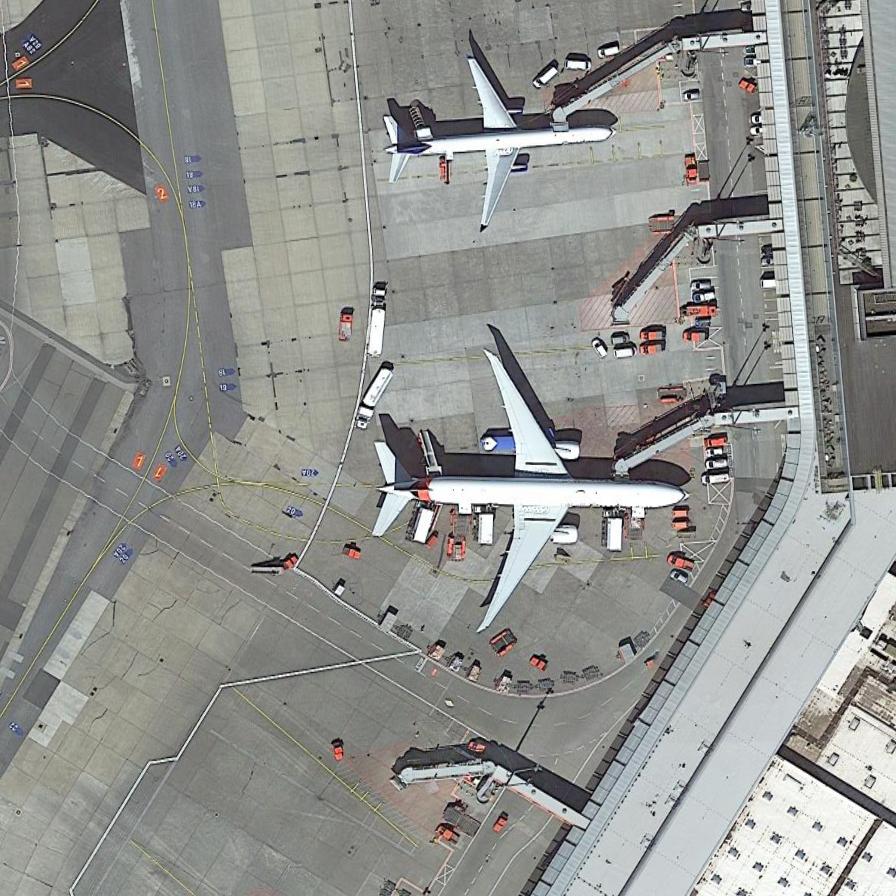}\vspace{4pt}
         \includegraphics[width=\linewidth]{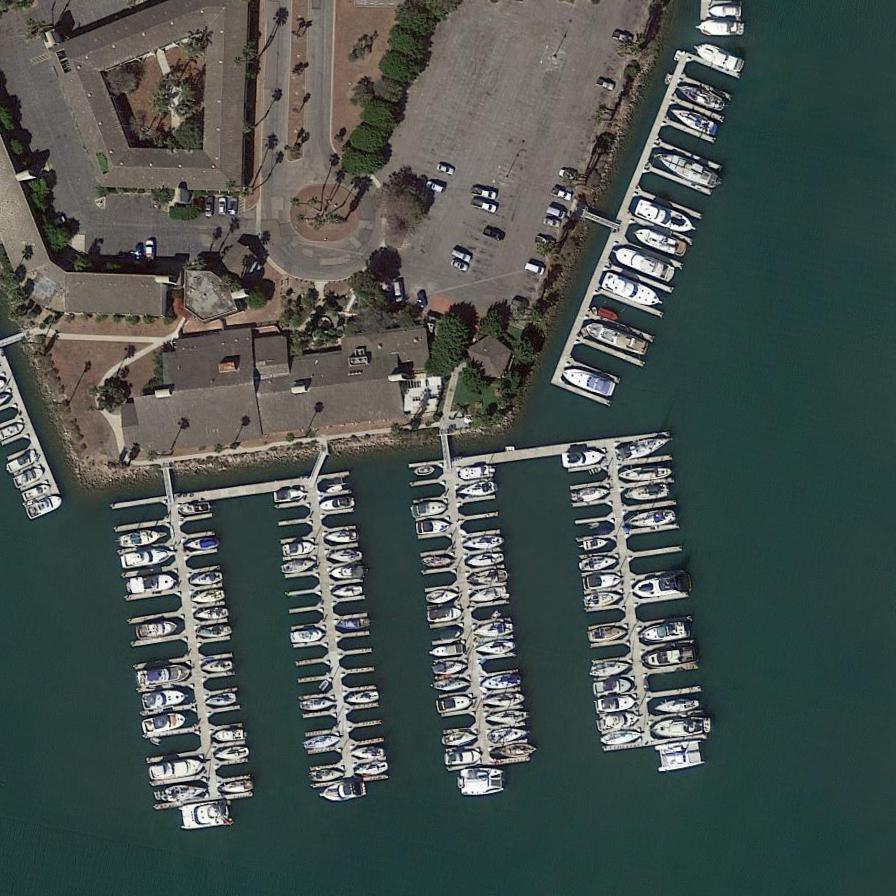}\vspace{4pt}
         \includegraphics[width=\linewidth]{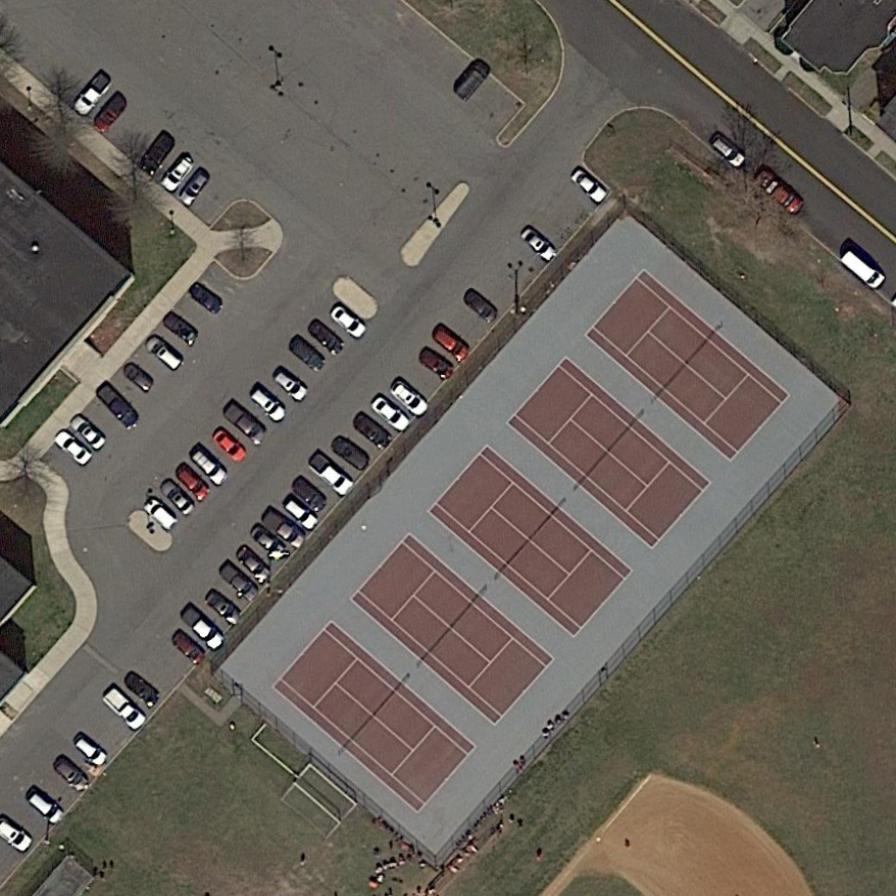}
      \end{minipage}
   }
   \subfigure[Prediction]{
      \begin{minipage}[b]{0.15\linewidth}
         \includegraphics[width=\linewidth]{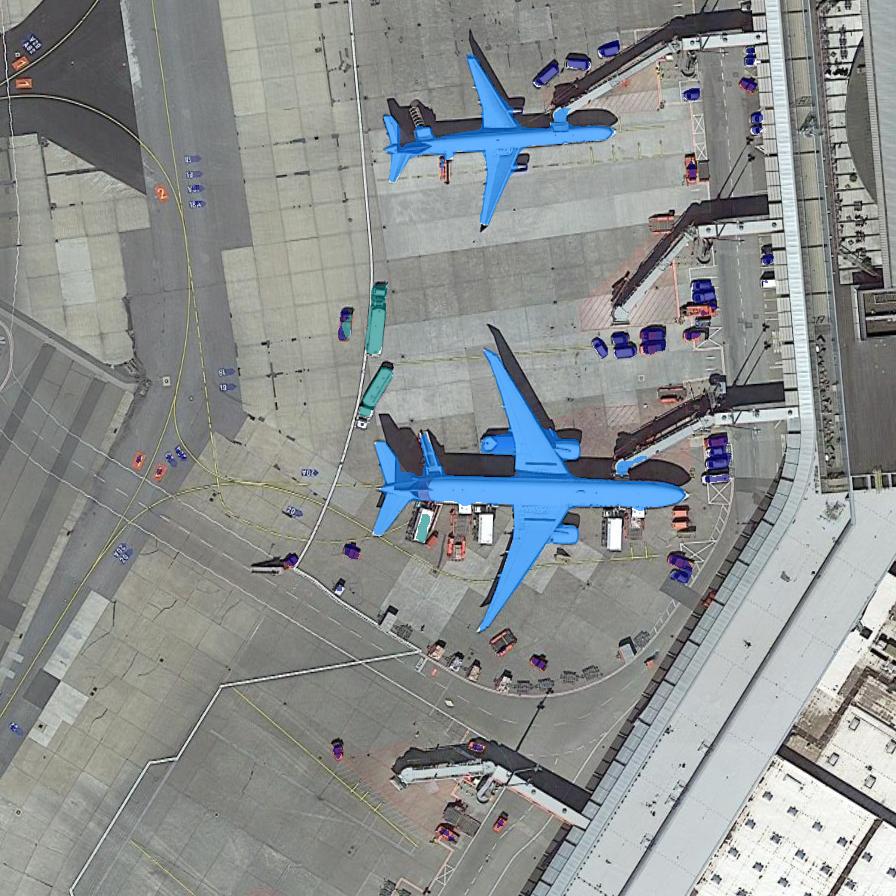}\vspace{4pt}
         \includegraphics[width=\linewidth]{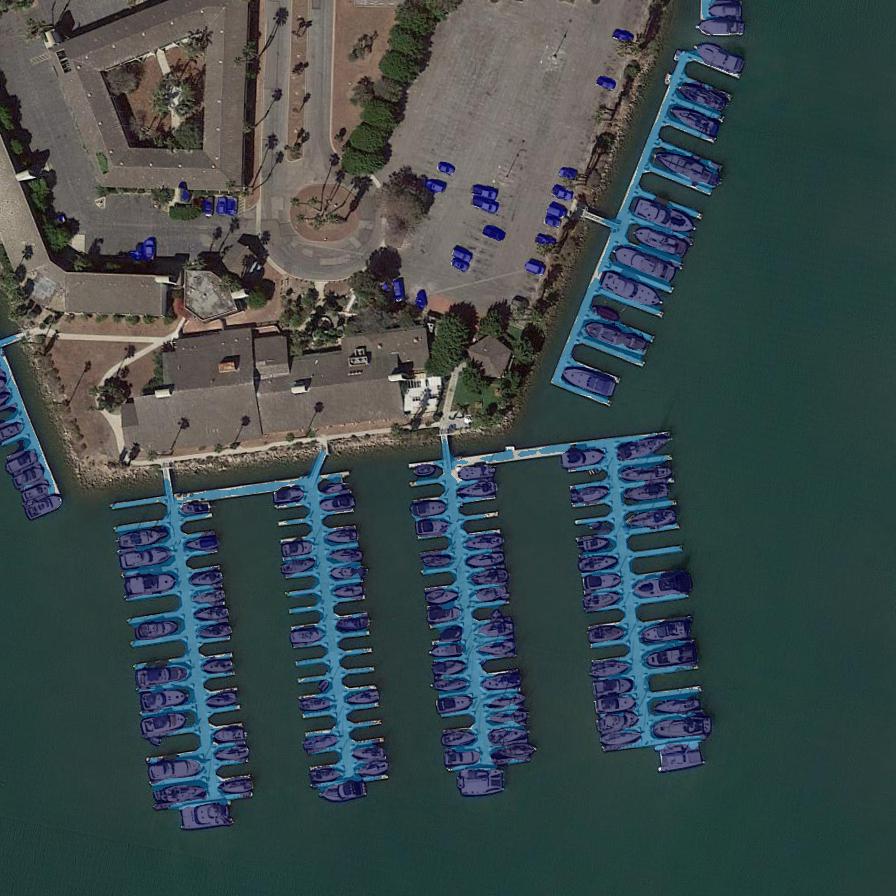}\vspace{4pt}
         \includegraphics[width=\linewidth]{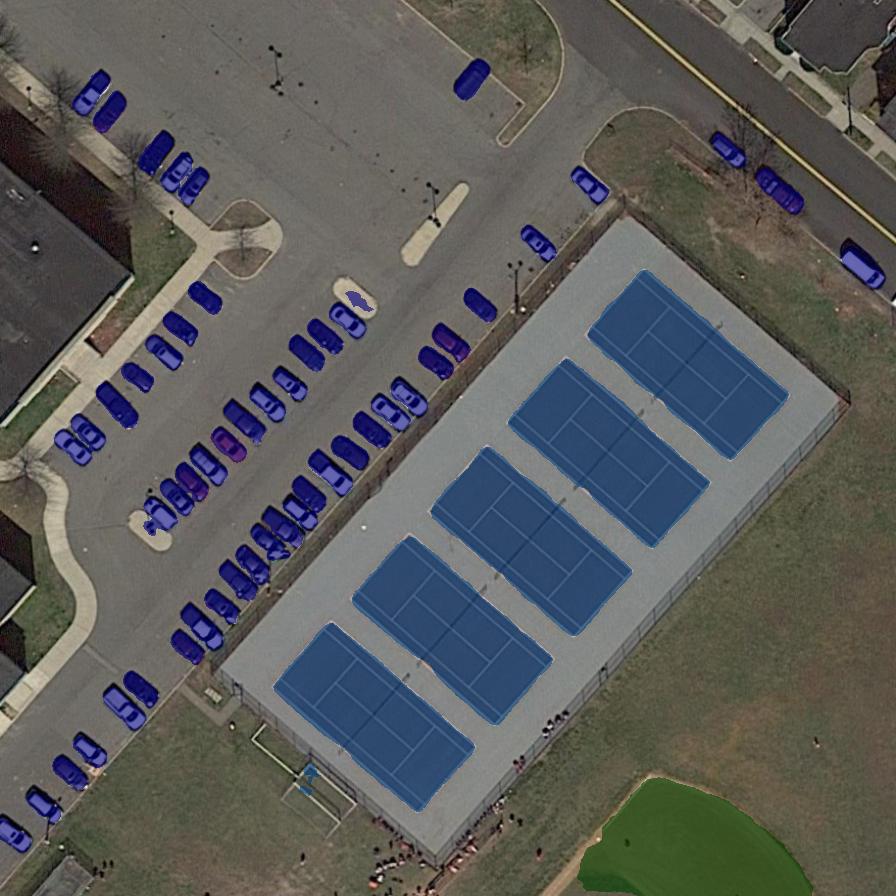}
      \end{minipage}
   }
   \subfigure[Relation $(OS = 4)$]{
      \begin{minipage}[b]{0.15\linewidth}
         \includegraphics[width=\linewidth]{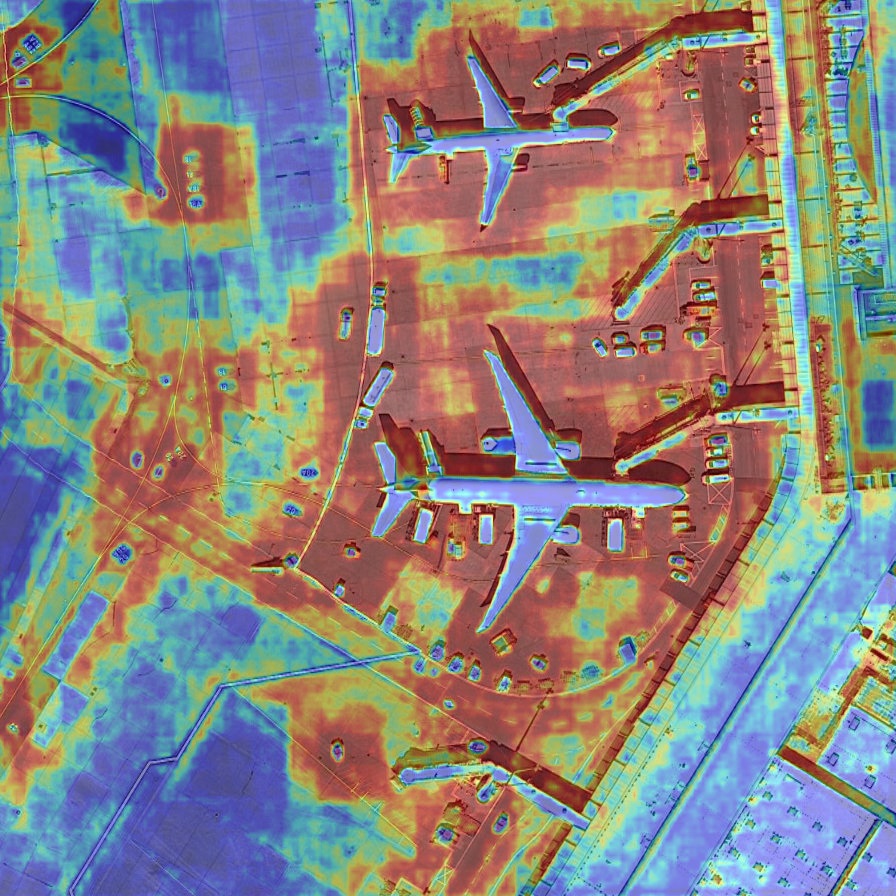}\vspace{4pt}
         \includegraphics[width=\linewidth]{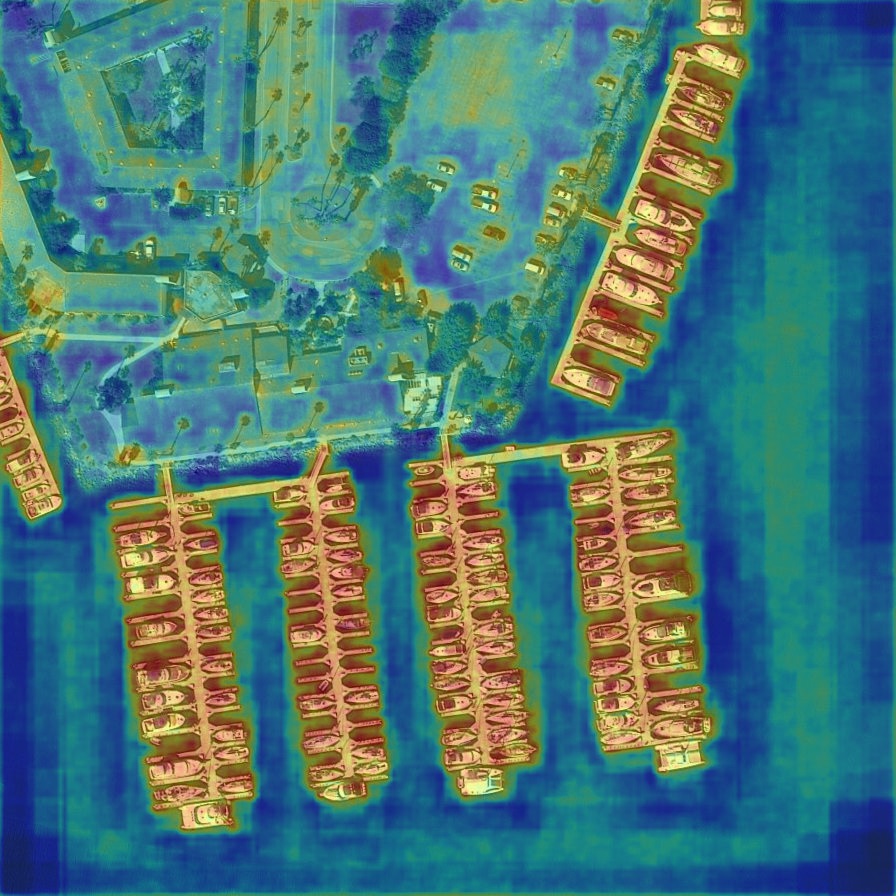}\vspace{4pt}
         \includegraphics[width=\linewidth]{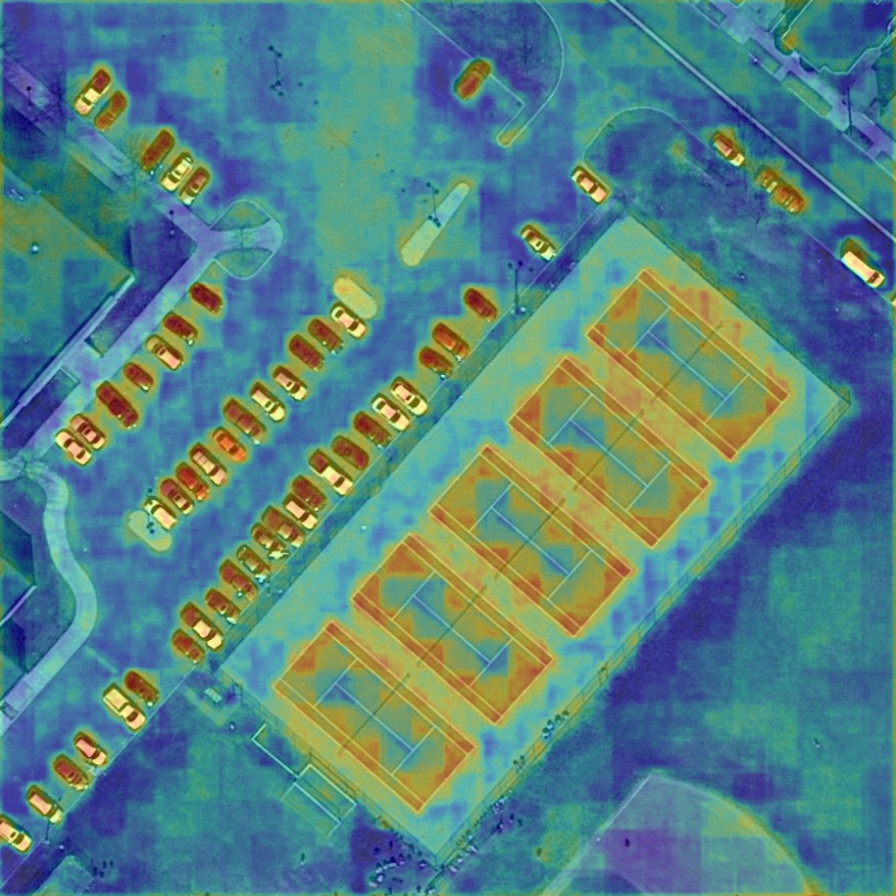}
      \end{minipage}
   }
   \subfigure[Relation $(OS = 8)$]{
      \begin{minipage}[b]{0.15\linewidth}
         \includegraphics[width=\linewidth]{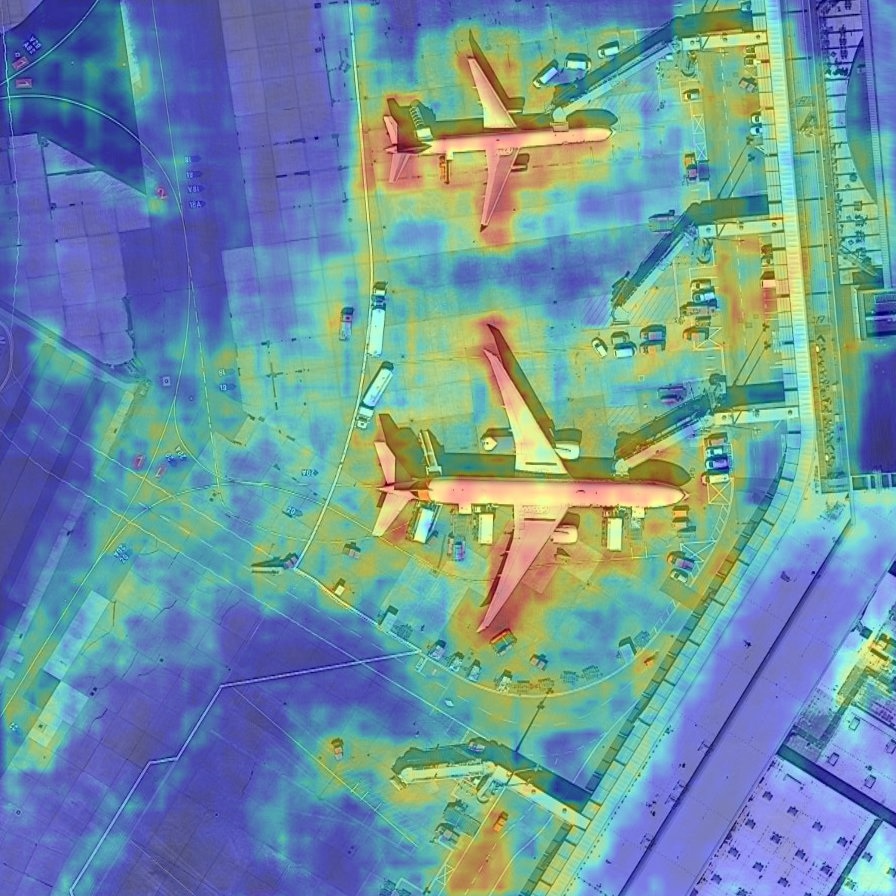}\vspace{4pt}
         \includegraphics[width=\linewidth]{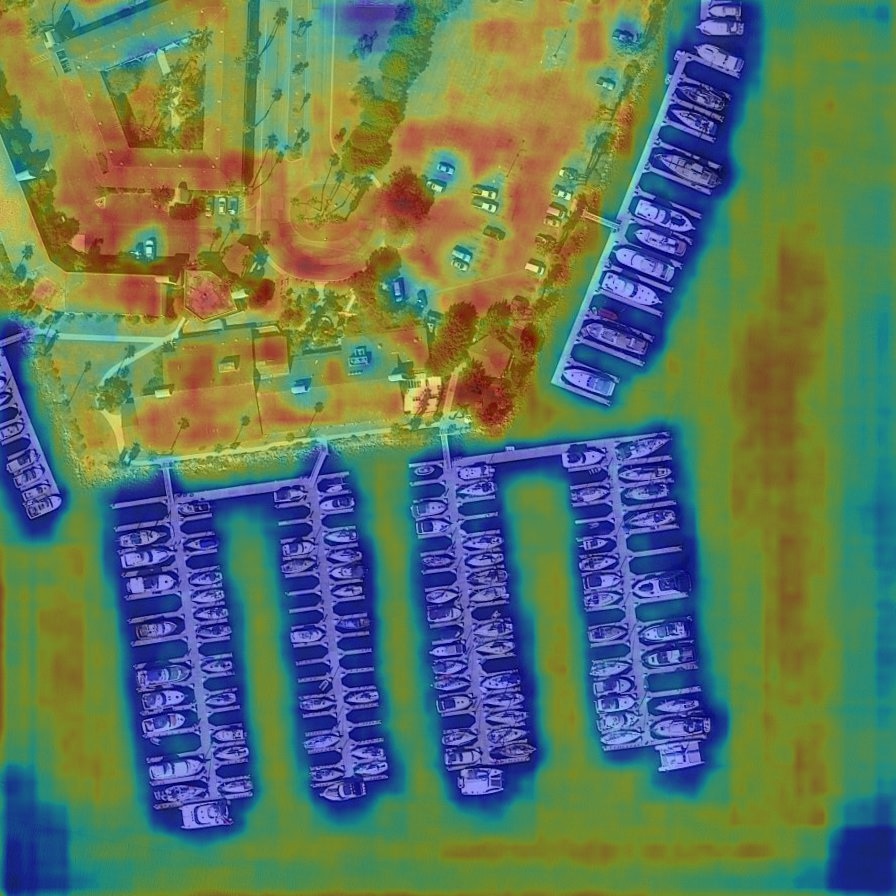}\vspace{4pt}
         \includegraphics[width=\linewidth]{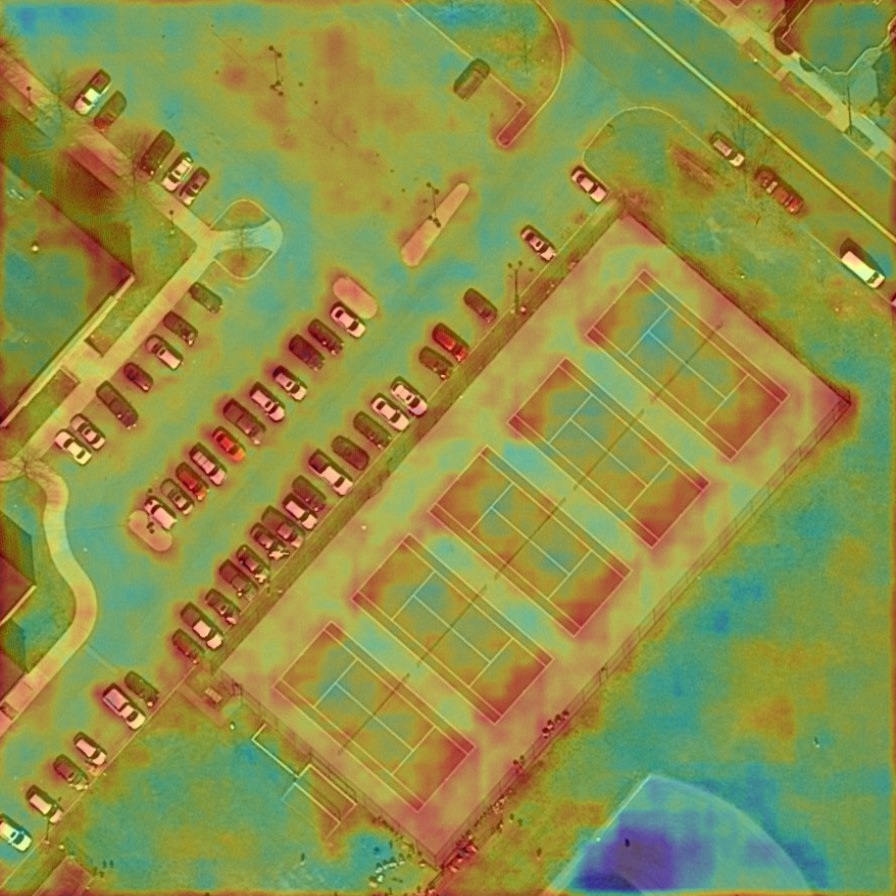}
      \end{minipage}
   }
   \subfigure[Relation $(OS = 16)$]{
      \begin{minipage}[b]{0.15\linewidth}
         \includegraphics[width=\linewidth]{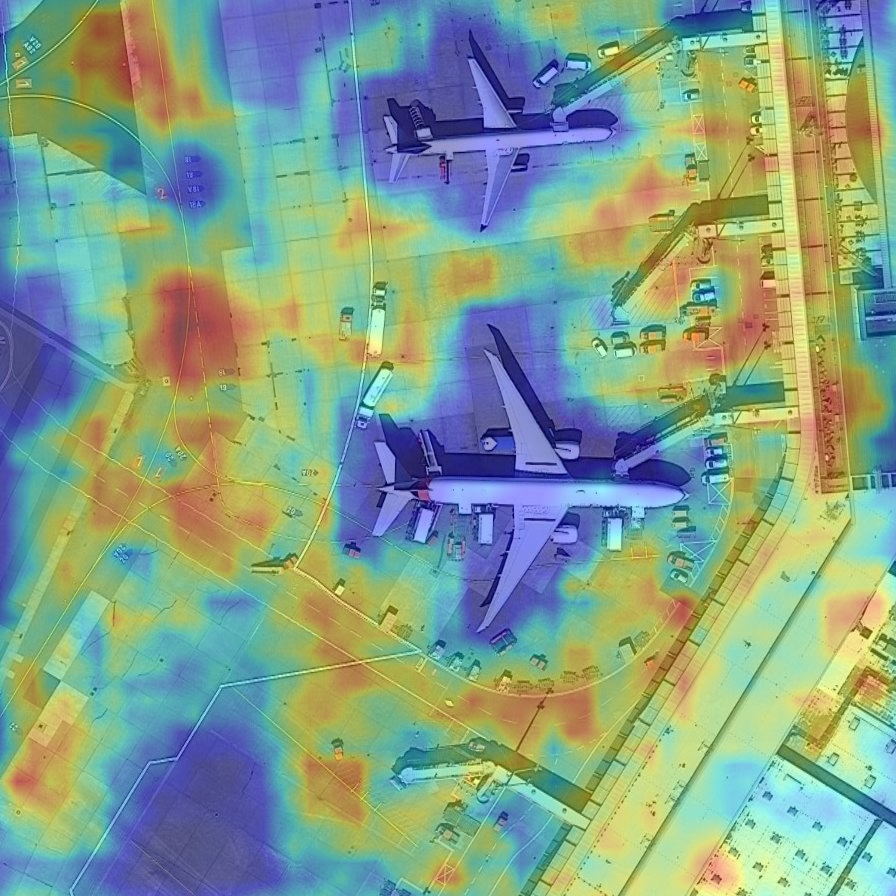}\vspace{4pt}
         \includegraphics[width=\linewidth]{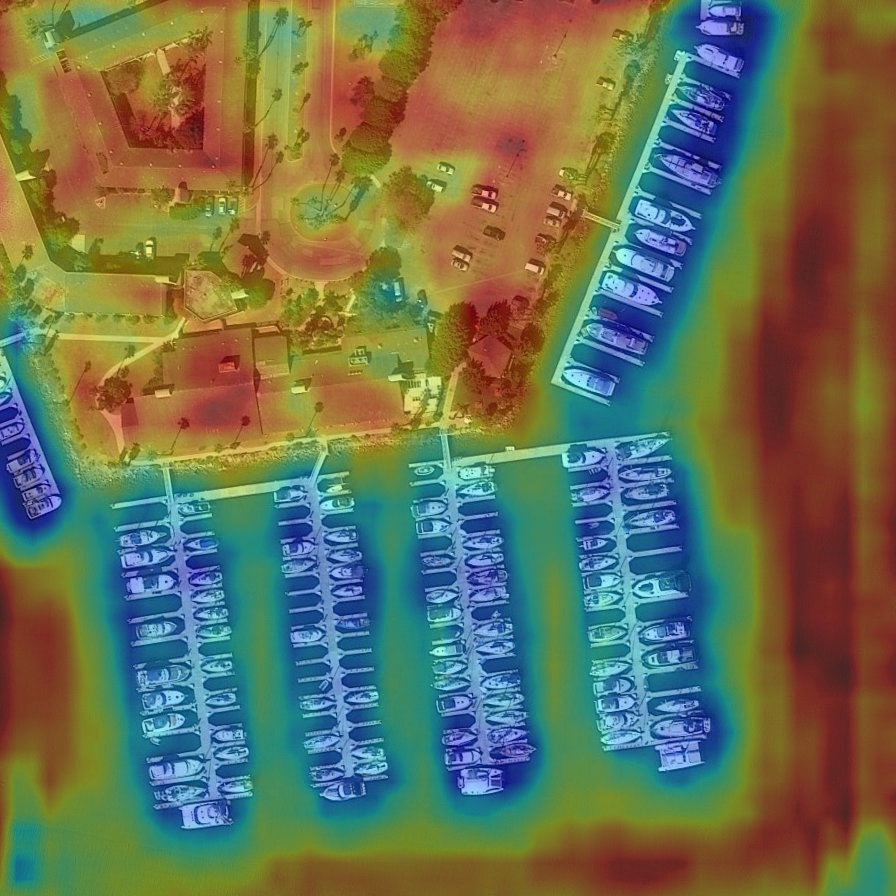}\vspace{4pt}
         \includegraphics[width=\linewidth]{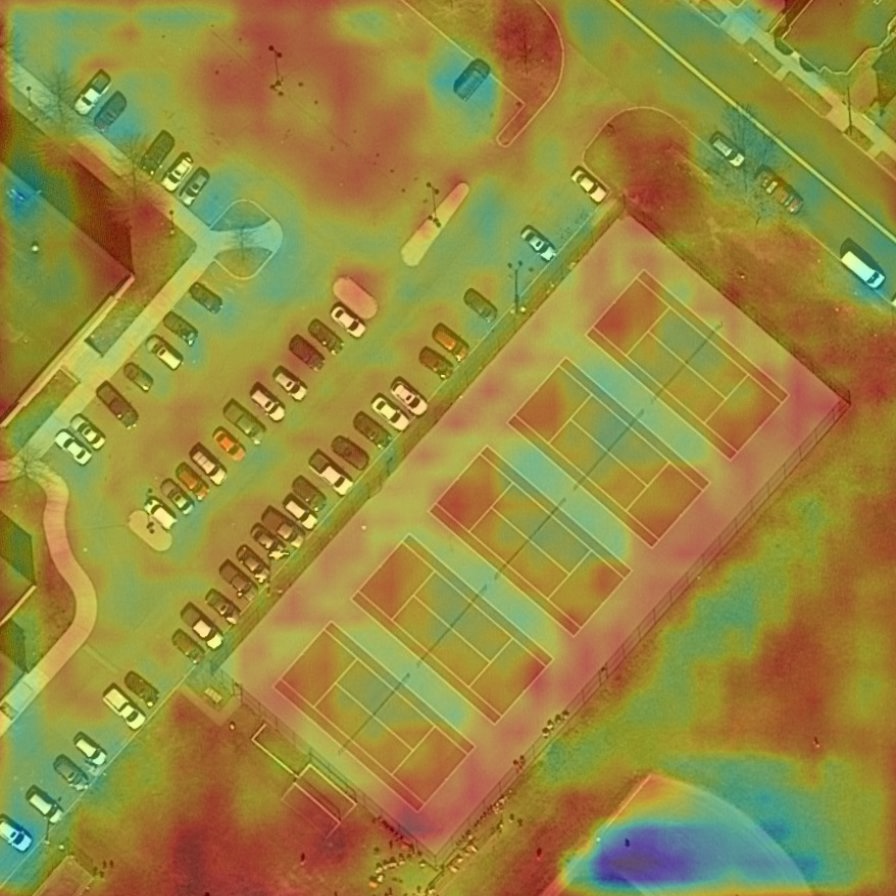}
      \end{minipage}
   }
   \subfigure[Relation $(OS = 32)$]{
      \begin{minipage}[b]{0.15\linewidth}
         \includegraphics[width=\linewidth]{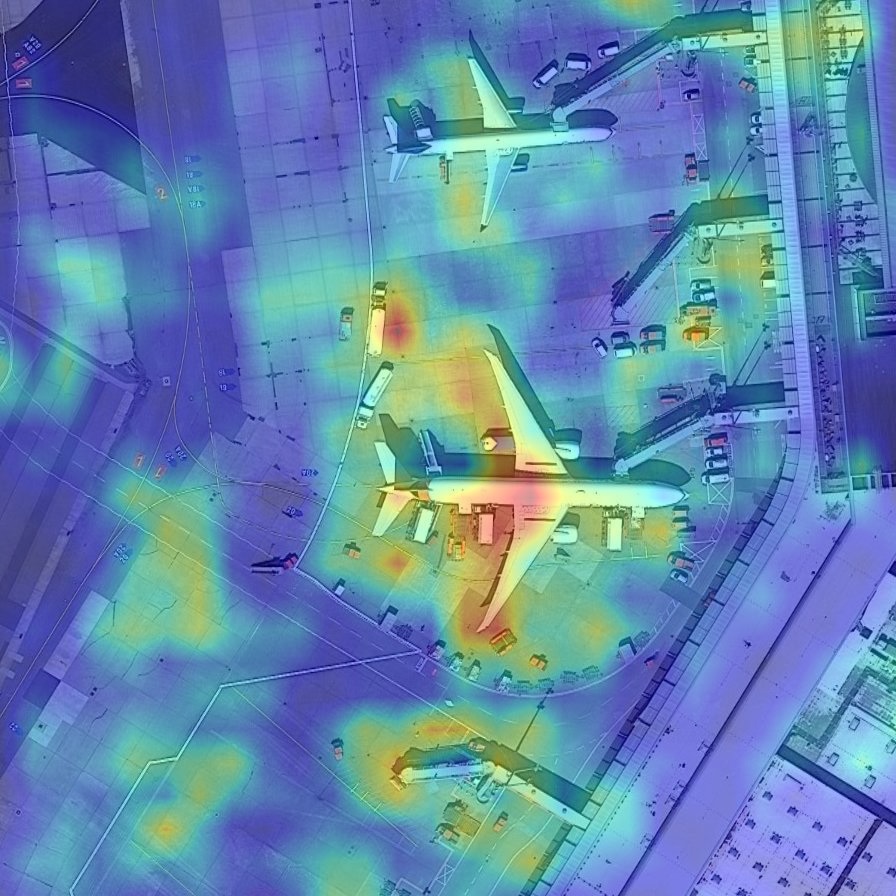}\vspace{4pt}
         \includegraphics[width=\linewidth]{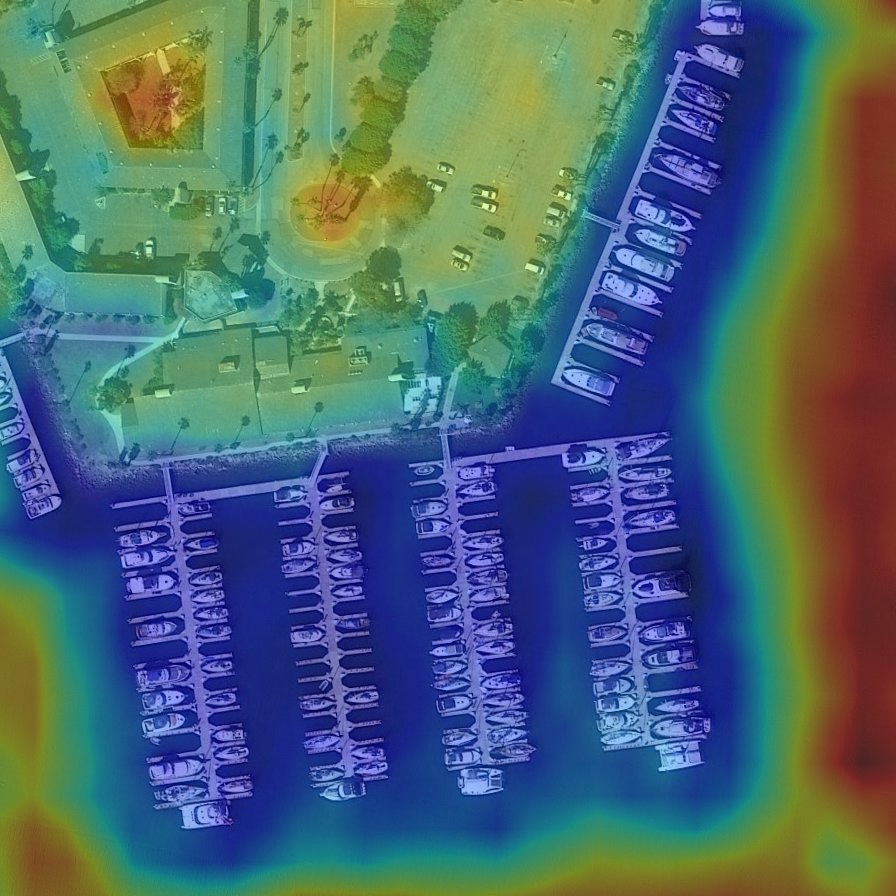}\vspace{4pt}
         \includegraphics[width=\linewidth]{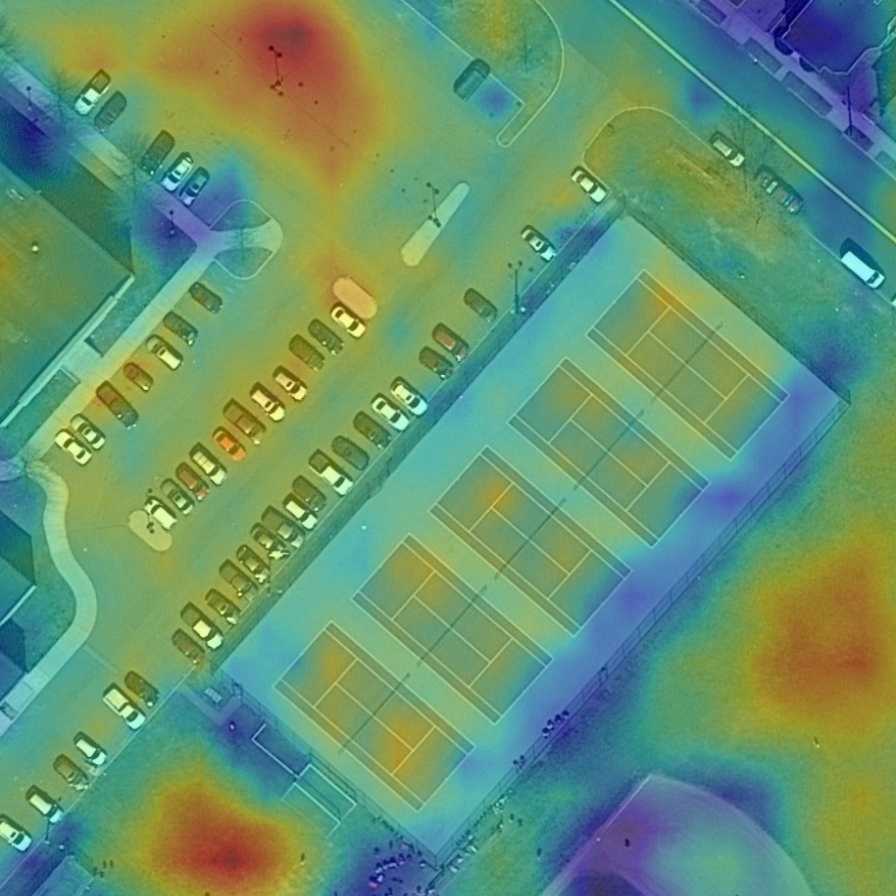}
      \end{minipage}
   }

   \caption{Visualization of F-S relation heatmap in the different pyramid levels. (a) original images. (b) object segmentation results. (c)-(f) images with F-S relation heatmaps in the different pyramid level. $OS$ denotes ``output stride'' defined in FPN. For convenient visualization, we resize these relation maps to corresponding image sizes.
   Legend: Scene 1 (\textcolor[rgb]{0,0.5,1}{plane}, \textcolor[rgb]{0,0.5,0.5}{large vehicle}, \textcolor[rgb]{0,0,0.5}{small vehicle}), 
   Scene 2 (\textcolor[rgb]{0,0,0.5}{small vehicle}, \textcolor[rgb]{0,0.39,0.608}{harbor}, \textcolor[rgb]{0,0,0.25}{ship}),
   Scene 3 (\textcolor[rgb]{0,0,0.5}{small vehicle}, 
   \textcolor[rgb]{0, 0.25, 0.5}{tennis court}, \textcolor[rgb]{0, 0.25, 0}{baseball diamond}), in a row order.}
   \label{fig:ablation_vis_relation}
\end{figure*}

\paragraph{Visual interpretation for F-S relation module.}
F-S relation module has good visual interpretability, combining with geoscience knowledge.
Fig.~\ref{fig:ablation_vis_relation} shows the visualization of F-S relations in the different pyramid levels.
Each pixel represents the relation intensity between the latent geospatial scene and the pixel-self.
There are three classical scenarios: airport, harbor, and parking-lot.
We can find that different scenarios focus on different objects that are discriminative to this scenario.
For example, the harbor mainly focuses on ship and water, while the airport focuses on the airplanes and their contexts.
Meanwhile, these relation maps illustrate again that the geospatial scene is related to scale, foreground, and foreground-relative contexts.
Because we can find that small objects are hot in the relation map with high spatial resolution ($OS =4$), such as small vehicle and ship.
The large objects are hot in the relation map with lower spatial resolution.
However, contexts are not spatial resolution-specific in the relation map.
It reveals that the geospatial scene is related to scale-specific foregrounds and scale-agnostic contexts.

\subsubsection{Foreground-Aware Optimization}
\paragraph{The effect of F-A optimization.}
Table~\ref{tab:module_ablation} (d) and (f) show the ablation results of adding F-A optimization based on baseline method (Table~\ref{tab:module_ablation} (a)) and baseline method with F-S relation and scale-aware projection (Table~\ref{tab:module_ablation} (c)), respectively.
F-A optimization boosts the performance with 2.2\% and 3.24\% in mIoU without any extra computation and memory footprint.
It indicates that F-A optimization can significantly alleviate foreground-background imbalance problem for object segmentation in the HSR remote sensing imagery.
Meanwhile, it suggests that F-A optimization is compatible with the F-S relation module well.

\paragraph{Normalization.}
Normalization is designed to only adjust the loss distribution without change of sum for avoiding gradient vanishing.
Table~\ref{tab:opt_ablation} (c) shows the result of adding normalization on the naive softmax focal loss (Table~\ref{tab:opt_ablation} (b)).
Due to instability the naive softmax focal loss for object segmentation, the mIoU drops 4.05\%.
However, when adding the normalization, the performance obtains significant improvement with 2.49\% in mIoU.
Compared with naive softmax focal loss, it gains 6.54\% in mIoU.
It suggests that the tuning the loss distribution without change of sum is the key to alleviate foreground-background imbalance problem.

\paragraph{Annealing function.}
The annealing function is used in dynamic weighting stage of F-A optimization.
It aims to alleviate the training instability due to the wrong hard example estimation in the period of early training.
Table~\ref{tab:opt_ablation} (d), (e), and (f) show the results of applying three proposed annealing functions.
We can find that annealing-based dynamic weighting boosts the performance via reducing the wrong hard example estimation in the period of early training.
Intuitively, the cosine annealing function obtains the most significant gains of 0.63\% in mIoU.
Because the cosine annealing function has a slow descent rate at the start and end of training, which can stably adjust the loss distribution for healthy convergence, compared with linear annealing function and polynomial annealing function.

\paragraph{The choice of the focusing factor $\gamma$.}
The focusing factor $\gamma$ is introduced to adjust the weight of hard examples.
Larger $\gamma$, larger weight on hard examples.
Following \cite{lin2017focal}, we use varying $\gamma$ to conduct experiments.
The results are presented in Table~\ref{tab:gamma}.
As $\gamma$ increase, the performance obtains continually improvement.
With $\gamma=2$, F-A optimization yields 3.29\% in mIoU improvement over the baseline, achieving the best result of 63.71\% in mIoU.
However, with $\gamma=5$, the performance drops.
The possible reason is that noise labels are wrongly seen as hard examples, as mentioned in \cite{li2019gradient}.

\begin{table}[]
   \caption{mIoU (\%) on iSAID $val$ set using varying $\gamma$ for F-A optimization.
      \label{tab:gamma}}
   \centering
   \renewcommand{\arraystretch}{1.4}
   \resizebox{\linewidth}{!}{
      \begin{tabular}{c|cccccc}
         \hline
         $\gamma$  & 0     & 0.3   & 0.5   & 1     & 2              & 5     \\ \hline
         mIoU (\%) & 60.42 & 61.35 & 62.48 & 62.99 & \textbf{63.71} & 62.61 \\ \hline
      \end{tabular}}
\end{table}

\section{Conclusion}
\label{sec:conc}
In this work, we argue that false alarm and foreground-background imbalance problems are the bottlenecks of object segmentation in the HSR remote sensing imagery, while general semantic segmentation methods ignore it.
To alleviate these two problems, we propose foreground-aware relation network (FarSeg), which learns foreground-scene relation to enhance the foreground features for less false alarms and trains the network using a foreground-aware optimization in foreground-background balanced fashion.
The comprehensive experimental results show the effectiveness of FarSeg and a better trade-off between speed and accuracy.

{\small
   \bibliographystyle{ieee}
   \bibliography{farseg_arxiv}
}

\end{document}